\documentclass{article}

\PassOptionsToPackage{numbers, compress}{natbib}
 \usepackage[preprint]{neurips_2026}

\usepackage[utf8]{inputenc} 
\usepackage[T1]{fontenc}    
\usepackage{hyperref}       
\usepackage{url}            
\usepackage{booktabs}       
\usepackage{amsfonts}       
\usepackage{nicefrac}       
\usepackage{microtype}      
\usepackage{xcolor}         
\usepackage{graphicx}      
\usepackage{wrapfig}

\title{Auditing Multimodal LLM Raters: Central Tendency Bias in Clinical Ordinal Scoring}

\author{%
  Jiaqing Zhang\textsuperscript{1} \quad
  Sandeep Elluri\textsuperscript{2} \quad
  Bhanu Cherukuvada\textsuperscript{2} \quad
  Yonah Joffe\textsuperscript{3} \\
  \textbf{Jessica Sena\textsuperscript{4} \quad
  Miguel Contreras\textsuperscript{4} \quad
  Scott Siegel\textsuperscript{4} \quad
  Subhash Nerella\textsuperscript{4}} \\
  \textbf{Catherine E.\ Price\textsuperscript{3} \quad
  Parisa Rashidi\textsuperscript{4}} \\[6pt]
  \textsuperscript{1}Department of Electrical \& Computer Engineering \\
  \textsuperscript{2}Department of Computer and Information Science and Engineering \\
  \textsuperscript{3}Department of Clinical and Health Psychology \\
  \textsuperscript{4}Department of Biomedical Engineering \\[2pt]
  University of Florida, Gainesville, FL 32611 \\[4pt]
  \texttt{jiaqing.zhang@ufl.edu} \quad \texttt{parisa.rashidi@bme.ufl.edu}
}

\begin{document}

\maketitle

\begin{abstract}
    Multimodal large language models (LLMs) are increasingly explored as automated evaluators in clinical settings, yet their scoring behavior on ordinal clinical scales remains poorly understood.
    We benchmark three frontier LLM families against supervised deep learning models for scoring Clock Drawing Test (CDT) images on two public datasets using the Shulman rubric.
    While fully fine-tuned Vision Transformers achieve the best calibration (MAE~$0.52$, within-1 accuracy~$91\%$), zero-shot LLMs remain competitive on tolerance-based agreement (GPT-5 MAE~$0.67$, within-1 accuracy~$92\%$) despite higher absolute error.
    However, per-score analysis reveals that all three LLM families exhibit a pronounced \emph{central tendency effect} (systematic endpoint compression): predictions are systematically compressed toward the middle of the scale, with over-prediction at the low end (score~$0 \to 1$) and under-prediction at the high end (score~$5 \to 4$).
    This effect disproportionately affects the clinically critical extremes where accurate scoring most impacts screening decisions for cognitive impairment.
    Targeted ablations show that neither few-shot exemplars spanning the full score range nor removing clinical terminology from the prompt eliminates the effect.
    Our findings extend the LLM-as-a-judge bias literature from NLP evaluation to clinical assessment, and highlight the need for calibration-aware evaluation and post-hoc calibration before deploying LLM-based raters in high-stakes screening workflows.
\end{abstract}

\section{Introduction}

Large language models (LLMs) have rapidly moved beyond text generation into the role of automated evaluators, a paradigm known as LLM-as-a-judge~\cite{gu2024survey}. 
Recent work~\cite{li2024llms, zheng2023judging} has demonstrated that LLMs can assess open-ended responses, rank competing outputs, and approximate human agreement on subjective rubrics, motivating their use as scalable proxies for expert annotation across natural language processing (NLP) benchmarks, educational assessment, and content moderation. 
However, this convenience comes with systematic distortions: position bias, verbosity bias, self-preference effects, and score-distribution anomalies have all been documented in structured evaluation settings~\cite{chen2024humans,li2025evaluating,ye2024justice}.
As multimodal LLMs are increasingly piloted for clinical tasks such as radiology interpretation and diagnostic scoring~\cite{wu2025exploring,wang2024beyond}, an urgent question arises: do these scoring biases persist, or worsen, when LLMs operate on clinical ordinal scales where prediction errors carry direct patient-facing consequences?

Standard practice for evaluating LLM raters relies on aggregate metrics: MAE, exact-match accuracy, within-tolerance agreement, borrowed from the NLP evaluation literature. We argue that these metrics can systematically conceal failure modes that matter clinically: a rater that achieves strong aggregate agreement while systematically failing at the scale extremes may pass conventional evaluation while being clinically unsuitable. We therefore propose an audit protocol combining per-score error decomposition, calibration-slope analysis, and a prompt-ablation suite designed to distinguish prompt-engineering artifacts from intrinsic model behavior.

We demonstrate the protocol on the Clock Drawing Test (CDT), an ideal setting to investigate this question.
CDT is a brief, widely administered bedside task in which patients draw a clock face showing a specified time; trained clinicians then score the drawing on an ordinal scale that captures visuospatial and executive function~\cite{shulman1993clock,zhang2024developing}.
Because scoring depends on the holistic interpretation of hand-drawn imagery and is subject to nontrivial inter-rater variability, both computer vision pipelines and multimodal LLMs have been proposed to automate the process~\cite{jimenez2023using,wysokinski2026screening,yang2026early}.
Yet when an LLM serves not as a language judge but as a \emph{clinical rater} of patient-produced drawings, the central issue shifts from overall accuracy to whether its scoring behavior is well calibrated and free of the systematic bias that could compromise downstream screening decisions.

In this work, we conduct an evaluation study of LLM-based CDT scoring and compare it with traditional deep learning (DL) approaches. 
Our principal finding is that all LLMs exhibit a pronounced \emph{central tendency effect}: predictions are systematically compressed toward the middle of the score range, overestimating poor drawings and underestimating strong ones.
This effect is consistent across models and persists under both few-shot prompting and de-clinicalized prompt variants, suggesting that it may reflect a broader calibration limitation of current multimodal LLM raters.

Our contributions are as follows:
\begin{itemize}
\item We propose a systematic evaluation protocol for evaluation of multimodal LLMs in clinical ordinal scoring tasks using CDT scoring, benchmarking three commercial LLM families against established deep learning approaches on the NHATS dataset~\cite{freedman2019cohort} with external validation on an independent CDT cohort~\cite{raksasat2023attentive}, and characterizing their comparative strengths and failure modes.
\item We demonstrate the central tendency effect in LLM-based clinical scoring: predictions are compressed toward the middle of the score range across all models tested, with disproportionate errors at the extremes of the scoring scale.
\item We isolate this bias from alternative explanations through targeted ablations: neither few-shot exemplars spanning the full score range nor removal of clinical terminology from the prompt eliminates the effect, suggesting it reflects intrinsic model behavior rather than a prompt-engineering artifact.
\end{itemize}

Although our empirical study focuses on cognitive impairment screening through the CDT, we believe the findings speak to a broader challenge. The central tendency effect we document, where LLM evaluators systematically avoid extreme scores, may be a general property of LLM-based assessment in structured scoring settings, particularly in contexts where asymmetric errors at the tails of the score distribution carry disproportionate consequences. We therefore view this work as both a contribution to automated clinical assessment and a step toward understanding the calibration properties of LLM-based evaluators more broadly.

\section{Related Work}
\label{sec:related}

\subsection{Automated CDT Scoring and Multimodal LLMs in Clinical Assessment}
\label{sec:rw-cdt}

The CDT is valued for its brevity and sensitivity to visuospatial and executive dysfunction~\cite{shulman2000clock,freedman1994clock}, yet manual neuropsychological scoring remains subjective: multiple systems exist, inter-rater agreement varies across settings, and the process does not scale to population-level screening~\cite{pinto2009literature}.
These limitations have motivated a substantial body of work on automated CDT scoring.
Early approaches used hand-crafted features (contour geometry, digit placement, hand angles) with classical classifiers~\cite{souillard2016learning}; subsequent deep learning methods moved to end-to-end convolutional pipelines, with CNN architectures achieving screening accuracies above 96\% on clinical cohorts~\cite{chen2020automatic,sato2022automated}.
More recent work has explored Vision Transformers~\cite{jimenez2024comparative} (ViT) and a self-supervised relevance-factor Variational Autoencoder (RF-VAE)~\cite{zhang2024developing} for clock drawing understanding.

In parallel, multimodal LLMs piloted across clinical domains: radiology case interpretation~\cite{wu2025exploring,yu2025pet}, histological grading~\cite{ono2024evaluating}, and broader diagnostic tasks~\cite{wang2024beyond}, with models approaching expert-level performance on structured examinations but exhibiting weaknesses in fine-grained image interpretation and calibration.
A small number of studies have begun applying multimodal LLMs to neuropsychological assessment, including CDT scoring~\cite{wysokinski2026screening} and speech-based cognitive screening~\cite{kashyap2025predicting}.


\subsection{Scoring Bias in LLM-based Evaluation}
\label{sec:rw-bias}

The use of LLMs as automated evaluators, commonly termed as \emph{LLM-as-a-judge}, has gained traction as a scalable alternative to human annotation for tasks where traditional metrics fail to capture semantic quality~\cite{zheng2023judging}.
However, a growing body of work has revealed systematic biases in LLM judges: position bias (preferring responses based on ordinal placement)~\cite{shi2025judging}, verbosity bias (favouring longer outputs regardless of quality)~\cite{zheng2023judging}, self-preference bias (assigning higher scores to the model's own outputs)~\cite{panickssery2024llm}, and significant sensitivity to surface-level prompt variations such as rubric order and score identifiers~\cite{li2025evaluating}.
Chen et al.~\cite{chen2024humans} compared human and LLM judgment biases, finding shared susceptibility to authority bias but divergent behavior on misinformation oversight.

Of particular relevance to our work, several studies have noted that LLM evaluators tend to compress their scoring distributions toward the center of the scale, avoiding extreme ratings~\cite{li2025evaluating,ye2024justice}.
Yet these findings derive almost exclusively from NLP evaluation settings: text summarization, dialogue quality, and instruction following, where miscalibration is a methodological concern but not a patient-safety issue.

No prior work has, to our knowledge, systematically examined whether central tendency effect manifests when LLMs serve as \emph{clinical raters} on ordinal scales, nor whether it can be mitigated through prompt design.
Our study addresses this gap: we quantify the effect in a controlled clinical scoring task, isolate it from confounding explanations through targeted ablations, and analyze its downstream impact on screening decisions where errors at the scale extremes carry disproportionate clinical consequences.

\section{Method}
\label{method}

\subsection{Dataset and Task Definition}

\paragraph{Dataset.}

We use clock-drawing images from two sources.
The primary dataset is drawn from the National Health and Aging Trends Study (NHATS)~\cite{freedman2019cohort}, a nationally representative longitudinal study of Medicare beneficiaries aged 65 and older, comprising 63{,}351 images across Rounds 1--13.
For external validation, we use 386 images from an independent public CDT cohort released with CDT-API-Network~\cite{raksasat2023attentive}, which contains paper-based clock drawings from a Thai clinical population.
In both datasets, participants draw an analog clock set to 11:10, and each drawing is scored on the Shulman six-level ordinal scale ($0$--$5$): $0$~=~not recognizable as a clock through $5$~=~accurate depiction~\cite{shulman1993clock}.
NHATS is used for model development and in-domain evaluation; the Thai cohort is reserved exclusively for external validation.

NHATS data are partitioned into development and test splits at an 80:20 ratio using participant-level stratification to prevent leakage from repeated longitudinal drawings.
For cross-paradigm comparison, we construct a \emph{score-balanced benchmark} of 597 images by sampling 100 images per score level from the NHATS test set (score~$0$ contributes all 97 available drawings).
This design ensures sufficient samples at the clinically critical extremes for reliable per-score error analysis.
All model families are evaluated on this identical set.

\paragraph{Task Definition.}

We cast CDT automation as \emph{ordinal clinical scoring} from image evidence.
Given a clock-drawing image~$x$, the goal is to predict an integer score $y\!\in\!\{0,1,2,3,4,5\}$ that matches the human-assigned reference label.
Two properties of this task distinguish it from standard image classification.
First, the labels are \emph{ordered}: the distance between predicted and true scores carries clinical meaning, so a one-step error (e.g., predicting~$4$ instead of~$5$) is far less consequential than a four-step error (e.g., predicting~$1$ instead of~$5$).
Second, the label distribution is \emph{imbalanced and concentrated at the extremes of clinical interest}: the lowest scores ($0$--$1$), which signal possible cognitive impairment, and the highest score ($5$), which indicates intact function, are precisely the categories where misclassification has the greatest downstream impact on screening decisions.
Any systematic tendency to under-predict extreme scores would therefore disproportionately affect the very cases that matter most for clinical triage.
 
To explore how different modeling paradigms interact with these properties, we compare three families of approaches:
\begin{enumerate}
    \item \textbf{Supervised convolutional learning (CNN):} learns hierarchical spatial features from pixel grids and maps them to discrete ordinal scores via a classification head.
    \item \textbf{Supervised token-based visual learning (ViT):} partitions the image into non-overlapping patches, models global dependencies through self-attention, and predicts either a discrete score or a continuous estimate $\tilde{y}\!\in\![0,5]$.
    \item \textbf{Rubric-driven multimodal reasoning (LLM-as-rater):} receives the drawing as a visual input together with a natural-language rubric describing the six score levels, and produces a score through in-context reasoning rather than gradient-based training on the target dataset.
\end{enumerate}
For each image, a model outputs either a discrete score~$\hat{y}$ (classification-based pipelines and LLMs) or a continuous estimate~$\tilde{y}$ (regression-based ViT variant), which is mapped to the integer $0$--$5$ axis via rounding for evaluation.

\subsection{Models}

All setups operate on normalized $224 \!\times\!224$ RGB inputs.
The CNN pipeline optionally applies a clock-extraction module (Otsu thresholding, morphological operations, connected-component cropping) to remove background clutter before learning.
Training-time augmentation includes horizontal flips, small rotations, and color jitter; inference uses deterministic resizing and ImageNet normalization.

\subsubsection{Deep Learning Models}

Our CNN baseline is a ResNet-101 pretrained on ImageNet.
Instead of a flat six-way softmax, we adopt cumulative ordinal modeling~\cite{frank2001simple} with five binary logits $\{z_k\}_{k=1}^{5}$, where $z_k$ represents the log-odds that the true score exceeds threshold $k{-}1$; the predicted score is $\hat{y}=\sum_{k=1}^{5}\mathbf{1}(\sigma(z_k)\geq 0.5)$
Training uses a weighted ordinal loss with tunable asymmetry and inverse-frequency sampling to counter class imbalance.

Both ViT variants replace the convolutional backbone with a pretrained Vision Transformer~\cite{dosovitskiy2020image}, whose patch-level self-attention provides a global receptive field that may better capture spatially distributed CDT cues (e.g., hand placement, digit spacing).
\textbf{ViT-Ordinal} reuses the cumulative-threshold head described above; model selection maximizes validation quadratic-weighted Cohen's~$\kappa$.
\textbf{ViT-Continuous} reframes scoring as bounded regression, predicting a scalar $\tilde{y}\!\in\![0,5]$ rounded to the nearest integer for evaluation; model selection minimizes validation MAE.
 
\subsubsection{Multimodal LLMs}
\label{sec:llm-models}
To represent the rubric-driven reasoning paradigm, we evaluate three state-of-the-art multimodal large language model families: GPT-5 \& GPT-5.4, Gemini-2.5-Pro, and Claude-4-Sonnet, each capable of accepting an image alongside a text prompt.
 
Unlike the supervised pipelines above, these models receive \emph{no gradient-based training} on NHATS clock images.
Instead, they are provided with a natural-language rubric that describes the six score levels and are asked to return an integer score for each drawing.
Because LLM-based scoring relies on in-context instruction following rather than learned decision boundaries, it offers a fundamentally different inductive bias: the model must \emph{interpret} visual evidence through linguistic clinical criteria.
The design and evaluation of the prompting strategies used to elicit scores are detailed in Section~\ref{sec:prompting}.

\subsection{Prompting Strategies}
\label{sec:prompting}
 
Each LLM receives the clock image together with an explicit $0$--$5$ scoring rubric and must return a structured JSON object containing the predicted score.
The rubric enumerates all six score levels, including both extremes ($0$: not recognizable as a clock; $5$: accurate depiction), so that the model has unambiguous anchors across the full ordinal range.
The inference prompts are fixed across all images and models. Full prompt text is provided in Appendix~\ref{app:prompts}.
 
All runs use deterministic decoding (temperature~$=0$, top-$p=1$) to minimize stochastic variance; output scores are validated and clamped to $[0,5]$ before evaluation.
 
\paragraph{Zero-shot vs.\ few-shot prompting.}
The default configuration for all three models is \emph{zero-shot}: the model receives only the rubric and the target image, with no scored examples.
To test whether explicit score anchoring can sharpen predictions at the boundaries of the scale, we additionally evaluate a \emph{few-shot} variant for GPT-5, in which 30 rubric-aligned exemplar images, 5 per score level, are prepended to the prompt.
By including exemplars that span the full $0$--$5$ range, this setup provides the model with concrete visual references for both extreme and intermediate scores, offering a direct test of whether in-context examples mitigate potential scoring conservatism.

\section{Experimental Setup}
\label{experiment}

\subsection{Data source}
All experiments are conducted on NHATS Clock Drawing Test (CDT) images with reference scores on a six-level ordinal scale from $0$ to $5$. 
Supervised models are trained using NHATS images and labels, whereas multimodal LLMs perform direct image scoring through prompting without gradient-based training on the target dataset. 
For cross-family comparison, all final results are reported on a shared held-out benchmark of 597 scored images.

\subsection{DL vs.\ multimodal-LLM comparison design}
We compare traditional deep learning (CNN, ViT-Ordinal, and ViT-Continuous) against multimodal LLM judges (GPT-5, GPT-5.4, Gemini-2.5-Pro, Claude-4-Sonnet) under the same CDT rubric.
All methods output scores on the same $0$--$5$ axis and are evaluated with identical downstream metrics.
Deep models are trained/fine-tuned on NHATS images, while LLMs receive the image and scoring rubric as prompt inputs and produce scores through direct multimodal inference.
Unless otherwise noted, LLM evaluation is zero-shot. A few-shot variant is additionally tested for GPT-5 as a targeted ablation of prompt-based score anchoring.

\subsection{Error-case analysis protocol}

We assess performance from three complementary perspectives. First, we measure absolute scoring error using mean absolute error (MAE) and root mean squared error (RMSE), which capture calibration quality on the ordinal scale. Second, we report within-1 accuracy, defined as the proportion of predictions within one score level of the reference label, to quantify tolerance-based agreement. Third, for comparability with prior CDT screening analyses, we report binary operating characteristics including sensitivity and specificity under a clinically motivated thresholding rule that maps ordinal CDT scores to screening categories (cognitive impaired when score~$<=3$). In addition to aggregate metrics, we examine per-score error patterns to characterize whether models systematically over- or under-predict particular regions of the scale.

\section{Results}

\subsection{Aggregate Comparison}\label{sec:aggregate}
Table~\ref{tab:main_results} summarizes performance across all model families on the 597-image CDT benchmark.
 
\paragraph{Supervised models.}
Among deep learning systems, ViT-Ordinal (unfrozen) achieves the strongest overall calibration, with an MAE of $0.52$, RMSE of $0.87$, and within-1 agreement ($91\%$).
ViT-Continuous (unfrozen) is the second-best supervised model (MAE~$0.65$, within-1~$89\%$), confirming that bounded regression is a viable alternative to ordinal classification, albeit with slightly coarser score resolution.
Frozen variants of both architectures perform substantially worse, underscoring the importance of end-to-end fine-tuning for this task.
 
\paragraph{Multimodal LLMs.}
Among zero-shot LLM judges, GPT-5 delivers the best score fidelity (MAE~$0.67$, within-1~$92\%$), followed by GPT-5.4 (MAE~$0.75$) and Gemini~2.5~Pro (MAE~$0.84$).
None of the LLMs outperform the fully fine-tuned ViT models on absolute calibration.
However, an intriguing pattern emerges when tolerance-based agreement is considered: GPT-5 achieves a within-1 accuracy of 92\%, comparable to ViT-Ordinal (unfrozen) (91\%; overlapping bootstrap CIs) despite a substantially higher MAE.
This suggests that GPT-5 often produces near-miss predictions that remain within one score level of the reference label, even when exact calibration is weaker.
This apparent paradox of competitive tolerance agreement alongside weaker exact calibration motivates the finer-grained analysis that follows.

\begin{table}[t]
  \caption{Results on the 597-image CDT benchmark with 95\% bootstrap confidence intervals (2,000 resamples).
           Lower MAE\,/\,RMSE is better; higher values are better for all other metrics.
           Best supervised result is \underline{underlined}; best LLM result is \textbf{bolded}.}
  \label{tab:main_results}
  \centering
  \small
  \begin{tabular}{lccccc}
    \toprule
    Model & MAE$\downarrow$ & RMSE$\downarrow$ & Within-1$\uparrow$ & Specificity$\uparrow$ & Sensitivity$\uparrow$ \\
    \midrule
    \multicolumn{6}{l}{\emph{Supervised visual models}} \\
    CNN ResNet-101
      & 1.46\,{\scriptsize[1.36,\,1.56]}
      & 1.90\,{\scriptsize[1.79,\,2.01]}
      & 0.61\,{\scriptsize[0.57,\,0.65]}
      & 0.19\,{\scriptsize[0.14,\,0.25]}
      & 0.86\,{\scriptsize[0.83,\,0.90]} \\
    ViT-Ord.\ (frozen)
      & 1.01\,{\scriptsize[0.94,\,1.09]}
      & 1.38\,{\scriptsize[1.30,\,1.46]}
      & 0.75\,{\scriptsize[0.71,\,0.78]}
      & 0.54\,{\scriptsize[0.47,\,0.60]}
      & 0.81\,{\scriptsize[0.77,\,0.85]} \\
    ViT-Ord.\ (unfrozen)
      & \underline{0.52\,{\scriptsize[0.47,\,0.58]}}
      & \underline{0.87\,{\scriptsize[0.80,\,0.94]}}
      & \underline{0.91\,{\scriptsize[0.89,\,0.93]}}
      & 0.85\,{\scriptsize[0.79,\,0.89]}
      & \underline{0.91\,{\scriptsize[0.89,\,0.94]}} \\
    ViT-Cont.\ (frozen)
      & 1.30\,{\scriptsize[1.22,\,1.37]}
      & 1.59\,{\scriptsize[1.51,\,1.67]}
      & 0.64\,{\scriptsize[0.60,\,0.68]}
      & \underline{0.94\,{\scriptsize[0.91,\,0.97]}}
      & 0.54\,{\scriptsize[0.50,\,0.59]} \\
    ViT-Cont.\ (unfrozen)
      & 0.65\,{\scriptsize[0.60,\,0.71]}
      & 0.93\,{\scriptsize[0.86,\,1.00]}
      & 0.89\,{\scriptsize[0.86,\,0.91]}
      & 0.89\,{\scriptsize[0.85,\,0.93]}
      & 0.89\,{\scriptsize[0.85,\,0.92]} \\
    \midrule
    \multicolumn{6}{l}{\emph{Multimodal LLM judges (zero-shot)}} \\
    GPT-5
      & 0.67\,{\scriptsize[0.62,\,0.73]}
      & 0.95\,{\scriptsize[0.89,\,1.01]}
      & 0.92\,{\scriptsize[0.89,\,0.94]}
      & 0.81\,{\scriptsize[0.75,\,0.86]}
      & 0.88\,{\scriptsize[0.85,\,0.91]} \\
    GPT-5.4
      & 0.75\,{\scriptsize[0.70,\,0.81]}
      & 1.02\,{\scriptsize[0.96,\,1.08]}
      & 0.89\,{\scriptsize[0.86,\,0.91]}
      & 0.45\,{\scriptsize[0.38,\,0.51]}
      & 0.97\,{\scriptsize[0.95,\,0.98]} \\
    Gemini 2.5 Pro
      & 0.84\,{\scriptsize[0.78,\,0.91]}
      & 1.17\,{\scriptsize[1.10,\,1.24]}
      & 0.82\,{\scriptsize[0.79,\,0.85]}
      & 0.28\,{\scriptsize[0.21,\,0.34]}
      & \textbf{0.99\,{\scriptsize[0.98,\,1.00]}} \\
    Claude 4 Sonnet
      & 0.87\,{\scriptsize[0.80,\,0.93]}
      & 1.18\,{\scriptsize[1.11,\,1.25]}
      & 0.81\,{\scriptsize[0.78,\,0.84]}
      & 0.31\,{\scriptsize[0.24,\,0.37]}
      & 0.96\,{\scriptsize[0.94,\,0.98]} \\
    \bottomrule
  \end{tabular}
\end{table}

\subsection{Per-Score Error Analysis}
\label{sec:per-score}

\begin{figure}[t]
  \centering
  \includegraphics[width=0.8\linewidth]{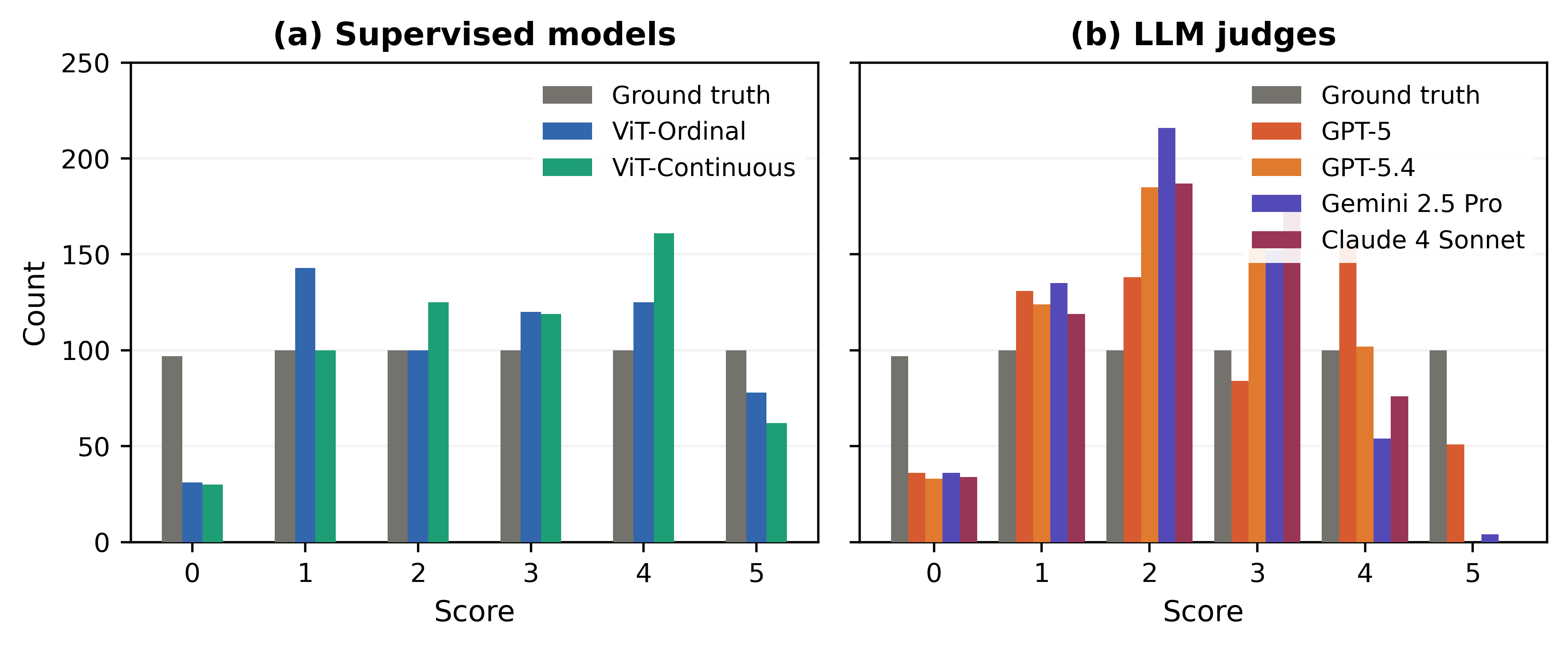}
  \caption{Predicted-score distributions versus ground truth.
           Supervised models (left) approximate the true label distribution;
           LLM judges (right) exhibit a compressed range with under-representation of extreme scores ($0$ and $5$) and over-representation of intermediate scores.
           Only models with MAE<1 are shown for clarity.}
  \label{fig:score-dist}
\end{figure}

The aggregate metrics in Table~\ref{tab:main_results} mask an important structural difference in \emph{where} each paradigm errs.
To expose this, we examine the predicted-score distributions and directional error profiles at each true score level.
 
\paragraph{Predicted-score distributions.}
Figure~\ref{fig:score-dist} overlays each model's predicted-score histogram on the ground-truth distribution.
 
Supervised models, particularly the unfrozen ViT variants, produce distributions that closely mirror the ground-truth histogram.
In contrast, all three LLMs generate markedly compressed distributions: scores~$0$ and~$5$ are substantially under-predicted, while intermediate scores, especially~$1$ and~$4$, are over-represented.
This compression accounts for the paradox noted in Section~\ref{sec:aggregate}: because LLM predictions cluster near the center of the scale, most errors are off by only one level, inflating within-1 agreement even as exact-match accuracy suffers.

\begin{wrapfigure}{r}{0.45\linewidth}
  \centering
  \includegraphics[width=\linewidth]{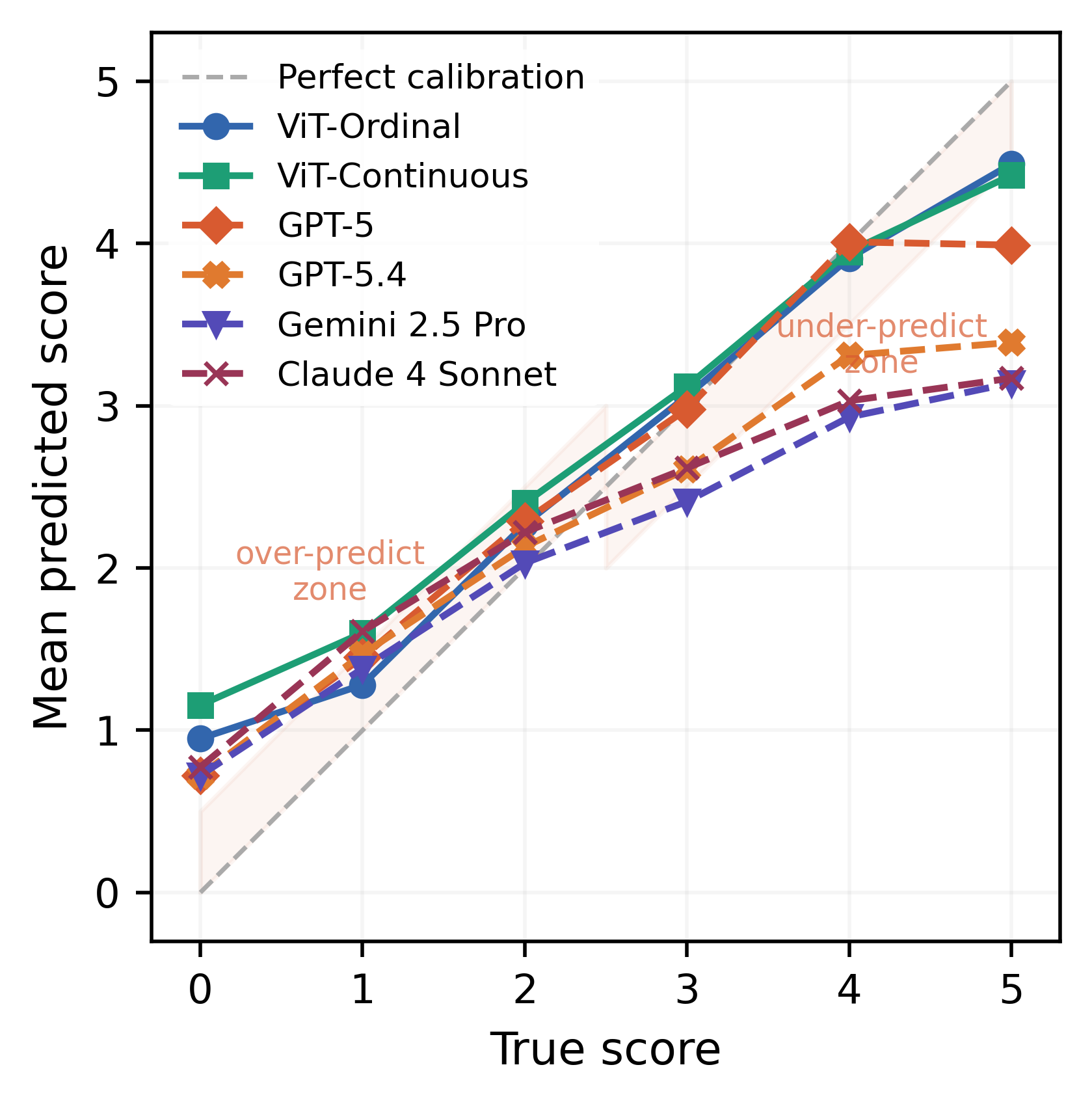}
  \caption{Score-level calibration. Supervised models (solid) cluster near the identity diagonal; LLM judges (dashed) exhibit shallower slopes.}
  \label{fig:calibration}
\end{wrapfigure}

\paragraph{Directional error profiles.}
Figure~\ref{fig:calibration} quantifies this compression by plotting the mean predicted score against the true score for each model.

Supervised models cluster near the identity diagonal, while all three LLMs produce calibration curves with noticeably shallower slopes: mean predictions lie above the diagonal at the low end (true scores~$0$--$1$) and below it at the high end (true scores~$4$--$5$).
A bootstrap test confirms that GPT-5's calibration slope is significantly lower than ViT-Ordinal's ($\Delta\hat{\beta} = -0.049$, 95\% CI $[-0.096, -0.002]$, $p = 0.020$), and a two-proportion $z$-test on toward-center error rates shows GPT-5 produces directionally biased errors at a significantly higher rate (34.0\% vs.\ 25.6\%, $z = 3.16$, $p < 0.001$).
The endpoint compression in LLM scoring is thus both statistically significant and structurally distinct from the error pattern of supervised models.

\begin{figure}[t]
  \centering
  \includegraphics[width=\linewidth]{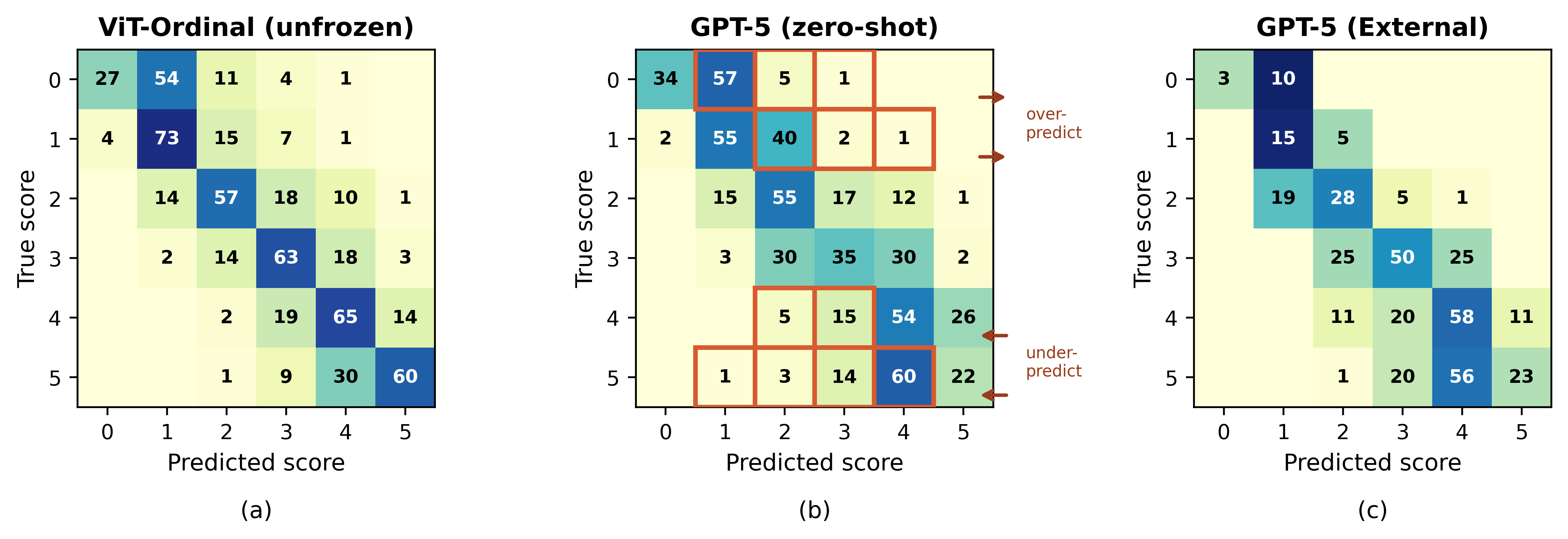}
  \caption{Confusion matrices for ViT-Ordinal (unfrozen) and GPT-5 (zero-shot) on NHATS and for GPT-5 (zero-shot) on the External Thai CDT cohort.
           Orange borders highlight towards-center errors.
           GPT-5's off-diagonal mass concentrates in the $(0{\to}1)$ and $(5{\to}4)$ cells, indicating systematic endpoint compression.}
  \label{fig:confusion}
\end{figure}
 
Figure~\ref{fig:confusion} (a) \& (b) reveals \emph{where} these errors concentrate at the individual-cell level.
ViT-Ordinal concentrates its mass tightly along the diagonal with no consistent directional pattern.
GPT-5 shows a markedly different structure: the $(0{\to}1)$ cell contains 57~samples, nearly $60\%$ of all true-$0$ drawings, while the $(5{\to}4)$ cell dominates with 60~samples, confirming symmetric compression from both ends.
In the interior of the scale (true scores~$2$--$3$), errors are smaller and more balanced in direction.
 
A revealing asymmetry strengthens this interpretation.
At true score~$4$, GPT-5 predicts score~$5$ in 26~cases, demonstrating that it is capable of assigning the maximum score.
Yet when the true score \emph{is}~$5$, GPT-5 assigns that score only 22~times, less often than it does for drawings that actually deserve a~$4$.
The same logic applies at the low end: true-$1$ drawings are almost never predicted as~$0$ (2~cases), yet true-$0$ drawings are frequently pulled up to~$1$ (57~cases).
This asymmetry indicates that the endpoint compression is not a perceptual limitation but a systematic scoring tendency: the model is capable of assigning extreme ratings in some contexts, yet assigns them substantially less often when they are the ground-truth labels.
Gemini~2.5~Pro and Claude~4~Sonnet exhibit the same toward-center structure with even greater severity (Appendix~\ref{app:confusion}).

\subsection{Prompt Ablations}

The central tendency effect documented above admits at least two alternative explanations that, if true, would reduce the finding to a prompt-engineering artifact rather than a behavioral property of LLMs.
We design two targeted ablations to test them.
 
\paragraph{Hypothesis~1: Insufficient score anchoring.}
Zero-shot prompts provide a textual rubric but no visual reference for each score level.
The model may default to ``safe'' middle scores simply because it lacks concrete exemplars of what a~$0$ or a~$5$ looks like.
If so, providing rubric-aligned exemplars should recalibrate the predicted distribution toward the extremes.
 
To test this, we evaluate LLMs with a few-shot prompt that prepends 30 scored exemplar images, five per score level, to the same rubric used in the zero-shot setting.
As shown in Table~\ref{tab:robustness}, few-shot prompting yields meaningful aggregate improvement: MAE drops from $0.67$ to $0.56$, and within-1 agreement rises from $92\%$ to $94\%$.
Accuracy at the top of the scale improves substantially (Acc$_{y=5}$: $22\% \to 52\%$), confirming that visual exemplars help anchor the high end.

However, the central tendency pattern is attenuated rather than resolved.
Even after few-shot prompting, nearly half of true-$5$ drawings are still scored below~$5$, and accuracy at the low end of the clinically more consequential extreme improves only modestly (Acc$_{y=0}$: $35.0\% \to 41.2\%$), meaning roughly $60\%$ of the most severely impaired cases are still over-predicted.
For comparison, ViT-Ordinal (unfrozen) achieves similar aggregate MAE ($0.52$) but without the systematic directional asymmetry at the endpoints: its errors are distributed across the full score range rather than concentrated at the extremes.
The predicted-score distribution under few-shot prompting remains visibly compressed relative to both the ground truth and the supervised models.
Explicit visual anchors, therefore, reduce the magnitude of central tendency effect but do not eliminate its characteristic structure endpoint compression with directional asymmetry.

The same pattern holds for Gemini~2.5~Pro and Claude~4~Sonnet under few-shot prompting (Confusion matrices in Appendix~\ref{app:confusion}).
 
\paragraph{Hypothesis~2: Safety-mechanism activation.}
Prior work has shown that multimodal LLMs can exhibit oversensitive or overly cautious behavior on benign inputs, and that such behavior may be amplified in medical contexts~\cite{li2024mossbench}.
It is therefore plausible that clinical terminology in our prompt keywords, such as ``neuropsychology'' and ``cognitive screening''activates a similar conservative mode, causing the model to avoid extreme judgments that might be perceived as consequential diagnostic statements.
If this mechanism underlies the central tendency effect documented in Section~\ref{sec:per-score}, then removing clinical framing from the prompt should attenuate endpoint compression.
 
To test this, we design a \emph{de-clinicalized} prompt variant that strips all medical and neuropsychological terminology, reframing the task as a generic image-quality evaluation.
The scoring scale, output format, and decoding parameters are otherwise identical to the clinical prompt (Section~\ref{sec:prompting}).
 
As shown in Table~\ref{tab:robustness} (row~3), removing clinical framing does not reduce central tendency effect; performance degrades substantially across all metrics (MAE: $0.67 \to 0.82$; RMSE: $0.95 \to 1.14$; within-1: $92\% \to 87\%$).
Per-score analysis confirms that the same directional error pattern, over-prediction at the low end, under-prediction at the high end, remains intact and is mildly amplified.
This result indicates that clinical language, rather than triggering conservative behavior, actually provides useful domain context that \emph{improves} scoring fidelity.
Removing it eliminates a potential source of anchoring without resolving the underlying scoring conservatism.

\begin{table}[t]
  \centering
  \caption{Robustness analysis: GPT-5 under three prompting conditions with 95\% bootstrap CIs.
           Per-score accuracy at the scale extremes is reported to track central tendency effect.}
  \label{tab:robustness}
  \small
  \begin{tabular}{lcccccc}
    \toprule
    Prompt condition & MAE$\downarrow$ & RMSE$\downarrow$ & W-1$\uparrow$ & Acc$_{y=0}$$\uparrow$ & Acc$_{y=5}$$\uparrow$ \\
    \midrule
    Zero-shot (clinical)
      & 0.67\,{\scriptsize[0.62,\,0.72]}
      & 0.95\,{\scriptsize[0.89,\,1.01]}
      & 0.92\,{\scriptsize[0.89,\,0.94]}
      & 0.35
      & 0.22 \\
    Few-shot (clinical)
      & 0.56\,{\scriptsize[0.51,\,0.60]}
      & 0.84\,{\scriptsize[0.79,\,0.90]}
      & 0.94\,{\scriptsize[0.92,\,0.95]}
      & 0.41
      & 0.52 \\
    Zero-shot (de-clin.)
      & 0.82\,{\scriptsize[0.76,\,0.89]}
      & 1.14\,{\scriptsize[1.06,\,1.22]}
      & 0.87\,{\scriptsize[0.84,\,0.89]}
      & 0.37
      & 0.05 \\
    \bottomrule
  \end{tabular}
\end{table}
 
\paragraph{Summary.} Neither enriching the prompt with scored exemplars nor removing clinical terminology eliminates the central tendency effect.
Few-shot prompting improves aggregate calibration, MAE drops by $17\%$, but the characteristic endpoint compression persists.
De-clinicalized prompting worsens all metrics, arguing against a purely clinical-framing explanation of the observed effect and revealing that clinical framing is beneficial rather than harmful.
Together, these results suggest that central tendency effect is not an artifact of prompt design but an intrinsic behavioral tendency of current multimodal LLMs when performing ordinal scoring tasks.

\subsection{External Replication on the Thai CDT Cohort}

External replication on the Thai CDT cohort further supports the central tendency pattern observed on NHATS. As shown in Figure~\ref{fig:confusion}(c), the confusion matrix exhibits the same qualitative error structure as in the in-domain setting: off-diagonal mass is concentrated in toward-center transitions, with low true scores tending to be over-predicted and high true scores tending to be under-predicted. In particular, endpoint errors are again asymmetric, indicating a tendency to under-assign extreme ratings. 

\section{Discussion}
\label{sec:discussion}

Our results illustrate the central claim motivating this work: aggregate agreement metrics can obscure a systematic failure mode of multimodal LLM raters in clinical ordinal scoring. Although GPT-5 achieves within-1 agreement comparable to the best supervised models, our evaluation protocol reveals a consistent deviation at both extremes toward the center of the scale: low scores are over-predicted, and high scores are under-predicted. In contrast, fully fine-tuned ViT models achieve stronger absolute calibration and do not exhibit the same endpoint asymmetry.

This distinction matters because errors at the extremes of the CDT scale are not clinically symmetric. Under-scoring severe impairment can reduce case detection, whereas downshifting intact drawings from 5 to 4 primarily increases false alarms and downstream review burden. Our findings, therefore, suggest that multimodal LLMs should not currently be used as standalone ordinal CDT raters when reliable identification of scale endpoints is important. A more plausible role is as a zero-shot prescreening tool or baseline, with calibrated supervised models or human raters handling final scoring.

More broadly, these findings extend the LLM-as-a-judge literature beyond NLP response evaluation to clinical absolute scoring.
Prior work has shown that LLM judges can align well with human preferences in some settings, but also exhibit systematic biases, including position bias and broader evaluation instability~\cite{zheng2023judging,shi2025judging,ye2024justice}.
Our results add a related failure mode in a patient-relevant ordinal task: endpoint compression under rubric-based scoring.
This pattern closely resembles central tendency effects long studied in human ordinal judgment~\cite{douven2018bayesian}, and we hypothesize that a similar mechanism operates in LLMs: alignment training via RLHF, which optimizes for human-preferred outputs, may internalize the same aversion to extreme ratings that human annotators exhibit, embedding it as a distributional prior in the model's scoring behavior.
Although class imbalance in pretraining corpora may also contribute to conservative prediction behavior, LLM scoring in our study is zero-shot with no exposure to the NHATS label distribution, and the directional asymmetry persists under both score-balanced evaluation and few-shot prompting with full-range exemplars, suggesting that imbalance alone does not explain the effect.
The prompt ablations further suggest that this effect is not easily removed by standard prompting interventions.

This study has several limitations. While we observe the same phenomenon on two independent CDT cohorts, both analyses remain within the Shulman-style scoring setting; broader validation across alternative rubrics and non-CDT clinical ordinal rating tasks remains future work. In addition, we study off-the-shelf LLMs with a limited set of prompt variants rather than adapted or calibrated models. A natural next step is to test whether lightweight post-hoc calibration or task-specific adaptation can reduce endpoint compression without sacrificing the flexibility of multimodal LLM raters \cite{guo2017calibration}.

\section{Conclusion}

We evaluated multimodal LLMs as raters for Clock Drawing Test scoring on a six-level clinical ordinal scale and compared them against supervised deep learning models on two independent CDT cohorts. Although frontier LLMs achieve strong tolerance-based agreement, per-score analysis reveals a consistent central tendency effect: predictions are compressed toward the middle of the scale, with over-prediction of low scores and under-prediction of high scores. These results show that aggregate agreement alone can mask clinically important scoring errors, and that evaluation of LLM-based raters should explicitly examine endpoint behavior and score-distribution compression. More broadly, our findings extend concerns about LLM-as-a-judge bias from NLP evaluation to clinical ordinal assessment, highlighting the need for calibration-aware evaluation before deploying multimodal LLMs in high-stakes screening workflows.

\begin{ack}
PC and PR were supported by the National Institute on Aging of the National Institutes of Health (NIH/NIA) under award number R56AG055337.
\end{ack}

\bibliographystyle{plainnat}
\medskip
\bibliography{references}


\appendix

\section{Technical Appendices and Supplementary Material}

\subsection{Supervised Model Details}

All supervised models share a common training protocol: two-phase optimization with frozen-backbone head alignment (Phase~1) followed by full fine-tuning at a reduced learning rate (Phase~2).
Input images are resized to $224\!\times\!224$ and normalized with ImageNet statistics.
Training uses fixed random seeds and deterministic CUDA settings for reproducibility.

\subsubsection{CNN (ResNet-101 Ordinal)}
\label{app:cnn}

\paragraph{Preprocessing.}
An optional clock-extraction stage reduces background clutter before augmentation: grayscale conversion, Otsu thresholding, morphological cleanup (erosion/dilation), connected-component selection by density and aspect ratio, and padded cropping.

\paragraph{Ordinal output design.}
The model predicts five cumulative logits $\{z_k\}_{k=1}^{5}$ corresponding to score thresholds.
The final score is decoded as $\hat{y}=\sum_{k=1}^{5}(\sigma(z_k)\geq 0.5)$, preserving label ordering.

\paragraph{Training.}
Phase~1: backbone frozen, head trained with SGD (lr $10^{-3}$, 15 epochs).
Phase~2: all layers unfrozen, SGD (backbone lr $10^{-5}$, head lr $10^{-3}$, 100 epochs) with momentum and weight decay.
Class imbalance is addressed via inverse-frequency weights and a weighted random sampler.

\begin{quote}
\small
\textbf{CNN training skeleton:}
\begin{verbatim}
model = ResNet101(pretrained=True)
head  = Linear(2048, 5)  # cumulative ordinal logits

phase1: freeze(backbone); train(head, lr=1e-3, epochs=15)
phase2: unfreeze(all); train(all, lr_bb=1e-5, lr_hd=1e-3, epochs=100)

pred_score = sum(sigmoid(z_k) >= 0.5 for k in [1..5])
\end{verbatim}
\end{quote}

\subsubsection{ViT-Ordinal}
\label{app:vit-ord}

Uses a pretrained ViT backbone with the same cumulative ordinal head as the CNN.
Phase~1: head only, AdamW (lr $10^{-3}$, 10 epochs).
Phase~2: full model, AdamW (lr $5\!\times\!10^{-6}$, 20 epochs, gradient clipping at 1.0).
Checkpoint selected by the best validation quadratic-weighted Cohen 's~$\kappa$.

\begin{quote}
\small
\textbf{ViT-Ordinal training skeleton:}
\begin{verbatim}
vit = ViT(pretrained=True)
ordinal_head = OrdinalHead(num_classes=6)

phase1: freeze(vit); train(ordinal_head, lr=1e-3, epochs=10)
phase2: unfreeze(vit); train(all, lr=5e-6, epochs=20, grad_clip=1.0)

select checkpoint by best validation weighted-kappa
\end{verbatim}
\end{quote}

\subsubsection{ViT-Continuous}
\label{app:vit-cont}

Shares the ViT backbone but replaces the ordinal head with a scalar regression head; output is clamped to $[0,5]$.
The training schedule and optimizer are identical to ViT-Ordinal.
The checkpoint selected by the best validation MAE.

\begin{quote}
\small
\textbf{ViT-Continuous training skeleton:}
\begin{verbatim}
vit = ViT(pretrained=True)
reg_head = Linear(hidden_dim, 1)

phase1: freeze(vit); train(reg_head, lr=1e-3, epochs=10)
phase2: unfreeze(vit); train(all, lr=5e-6, epochs=20, grad_clip=1.0)

pred_score = clip(output_scalar, 0, 5)
select checkpoint by best validation MAE
\end{verbatim}
\end{quote}

\subsection{Multimodal LLM Technical Details}
\label{app:llm}

Each API request includes a system prompt with NHATS-style score definitions ($0$--$5$), a user instruction enforcing JSON-only output, and the target image encoded as base64.
Inference uses deterministic decoding (temperature~$=0$, top-$p\!=\!1$).

\paragraph{Operational pipeline.}
Requests are dispatched with bounded parallelism and per-model rate throttling.
Transient API failures trigger up to three retries with exponential backoff.
Responses are parsed as JSON; if malformed, a regex-based fallback attempts to recover the score integer before marking the sample invalid.
All extracted scores are clamped to $[0,5]$.

\paragraph{Few-shot mode.}
The message sequence is extended with a class-balanced support bank of 5 labeled reference images per score level (scores $0$--$5$), prepended before the target image.
The output contract (single JSON score) is unchanged, enabling direct zero-shot vs.\ few-shot comparison.

\begin{quote}
\small
\textbf{LLM inference skeleton:}
\begin{verbatim}
messages = [
  {"role": "system",  "content": rubric_prompt},
  {"role": "user",    "content": json_instruction},
  {"role": "user",    "content": [fewshot_examples..., target_image]}
]
response = chat_completion(messages, temperature=0, top_p=1)
score = parse_json(response) ?? regex_fallback(response)
score = clamp(score, 0, 5)
\end{verbatim}
\end{quote}

\subsection{Reproducibility Notes}
\label{app:repro}

\paragraph{Hardware.}
All supervised model training and evaluation were performed on a single NVIDIA B200 GPU.
CNN (ResNet-101) training takes approximately \texttt{[X]} hours (15 frozen-head epochs + 100 fine-tuning epochs).
Each ViT variant trains in approximately \texttt{[X]} hours (10 + 20 epochs).
LLM inference was performed via commercial APIs and does not require local GPU resources.
 
\paragraph{Software and determinism.}
All training and evaluation scripts use Python 3.10 with PyTorch 2.7.
Fixed random seeds (\texttt{seed=42}) and deterministic CUDA settings (\texttt{torch.backends.cudnn.deterministic=True}) are applied throughout.
LLM API calls use deterministic decoding (temperature~$=0$, top-$p=1$); all calls were executed between 03/07/2026 and 05/04/2026.
 
\paragraph{Evaluation protocol.}
Model outputs across all paradigms are evaluated on the same score-balanced 597-image benchmark using identical metric definitions: MAE, RMSE, exact-score accuracy, within-1 agreement, specificity, and sensitivity.
95\% confidence intervals are computed via bootstrap resampling (2{,}000 iterations) over prediction--label pairs.
 
\paragraph{Code and data availability.}
Training scripts, evaluation pipelines, prompt templates, and figure-generation code are available at \url{[anonymous GitHub]}.
The NHATS clock-drawing repository is publicly accessible at \url{https://nhats.org}.
The external Thai CDT cohort is available through~\cite{raksasat2023attentive}.

\subsection{Full LLM Prompts}
\label{app:prompts}

\paragraph{Clinical prompt (default).}

\begin{quote}
\small
\textbf{System prompt:}
\begin{itemize}
\item You are a neuropsychology expert.
\item Score CDT images using NHATS criteria and published definitions.
\item Assign scores in the range $0$--$5$ only:
      0 (not recognizable as a clock),
      1 (severely distorted),
      2 (moderately distorted),
      3 (mildly distorted),
      4 (reasonably accurate),
      5 (accurate depiction of the clock task).
\item Do not assign negative or administrative codes.
\item Do not provide explanation or reasoning text.
\item Output must be valid JSON only: \texttt{\{"score": <0--5>\}}.
\end{itemize}
\textbf{User prompt:} ``Score this Clock Drawing Test image. Return ONLY the JSON object.''
\end{quote}

\paragraph{De-clinicalized prompt (ablation).}

\begin{quote}
\small
\textbf{System prompt:}
\begin{itemize}
\item You are an image quality assessment expert.
\item You are scoring a Clock Drawing image based on how accurately it depicts a clock showing the time 11:10.
\item Assign scores in the range $0$--$5$ only:
      0 (not recognizable as a clock),
      1 (severely distorted),
      2 (moderately distorted),
      3 (mildly distorted),
      4 (reasonably accurate),
      5 (accurate depiction of the clock task).
\item Do not assign negative or administrative codes.
\item Do not provide explanation or reasoning text.
\item Output must be valid JSON only: \texttt{\{"score": <0--5>\}}.
\end{itemize}
\textbf{User prompt:} ``Score this Clock Drawing Test image. Return ONLY the JSON object.''
\end{quote}

\subsection{Full Confusion Matrices}
\label{app:confusion}

\begin{figure}[h]
  \centering
  \includegraphics[width=\linewidth]{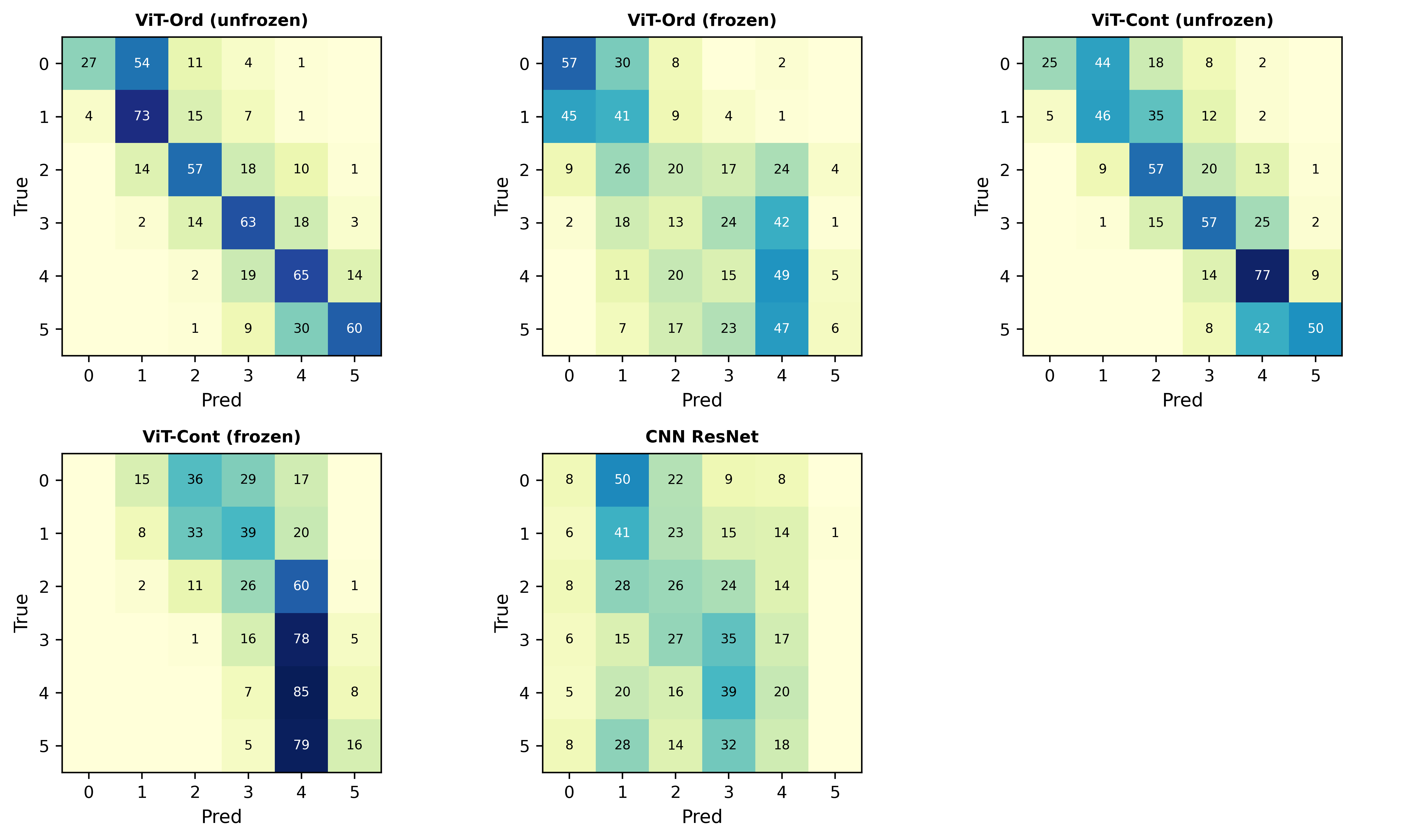}
  \caption{Confusion matrices for supervised models.
           Unfrozen ViT variants concentrate mass along the diagonal with no systematic directional pattern.
           Frozen variants and the CNN baseline show diffuse errors, but without the consistent toward-center asymmetry observed in LLMs.}
  \label{fig:cm-supervised}
\end{figure}

\begin{figure}[h]
  \centering
  \includegraphics[width=\linewidth]{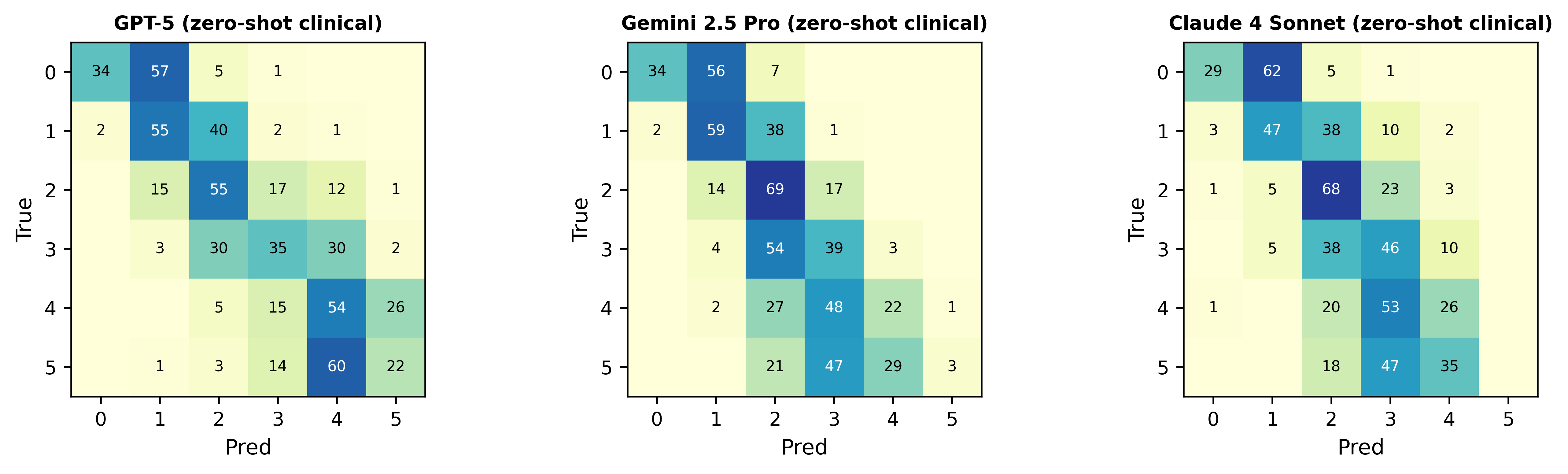}
  \caption{Confusion matrices for LLM judges under zero-shot clinical prompting.
           All three models exhibit the toward-center pattern: off-diagonal mass concentrates in the $(0{\to}1)$ and $(5{\to}4)$ cells.
           The effect is most extreme for Gemini~2.5~Pro, which predicts score~$5$ only 3 times out of 100 true-$5$ drawings.}
  \label{fig:cm-llm-zs}
\end{figure}

\begin{figure}[h]
  \centering
  \includegraphics[width=\linewidth]{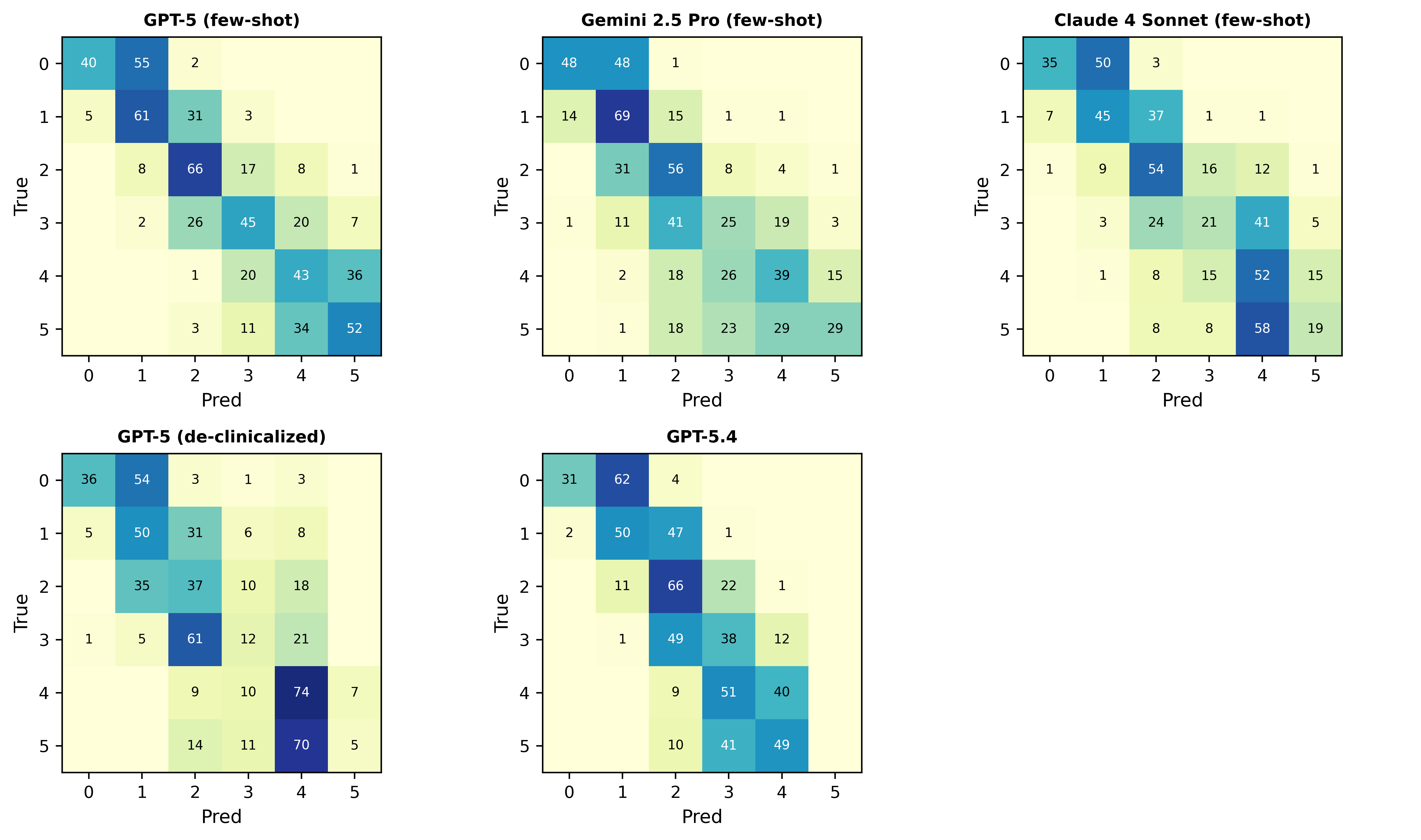}
  \caption{Confusion matrices under prompt ablations.
           Few-shot prompting increases diagonal mass most notably for GPT-5 at score~$5$ ($22 \to 52$) but the toward-center structure persists across all models and conditions.
           The de-clinicalised prompt (GPT-5) amplifies endpoint compression, with score-$5$ accuracy dropping to near zero.}
  \label{fig:cm-llm-ablation}
\end{figure}

We present the complete $6\!\times\!6$ confusion matrices for all evaluated models and prompt conditions on the 597-image score-balanced benchmark.
Matrices are organised into three groups: supervised models (Figure~\ref{fig:cm-supervised}), LLM judges under zero-shot clinical prompting (Figure~\ref{fig:cm-llm-zs}), and LLM judges under prompt ablations (Figure~\ref{fig:cm-llm-ablation}).
 
\newpage

\end{document}